\documentclass{article}

\usepackage[final,sglblindworkshop]{neurips_2026}
\workshoptitle{Continual Learning in the Era of Foundation Models and
Embodied Agents}

\usepackage[utf8]{inputenc}
\usepackage[T1]{fontenc}
\usepackage{hyperref}
\usepackage{url}
\usepackage{booktabs}
\usepackage{amsfonts}
\usepackage{nicefrac}
\usepackage{microtype}
\usepackage{xcolor}
\usepackage{graphicx}
\usepackage{multirow}

\title{Tracing the Thoughts of a Coding Agent Playing ARC-AGI-3: Lessons for Continual Learning}

\author{%
  Chen Wu \\
  AWS \\
  \texttt{wuc@amazon.com}
  \And
  Josh Passenger \\
  AWS \\
  \texttt{jospas@amazon.com}
  \AND
  Yin Song \\
  AWS \\
  \texttt{yinsong@amazon.com}
}

\begin{document}

\maketitle

\begin{abstract}
We study how a coding agent learns across a sequence of abstract reasoning tasks. The agent runs on
a frozen foundation model inside a fixed harness and acts by writing and running Python and shell
scripts. The agent retains no state across turns other than its written artifacts, so every
thought it forms, carries, corrects or abandons leaves a trace, where a thought is any belief,
rule or plan committed to a file. We let the agent play ARC-AGI-3, a set of interactive reasoning
games that provide no instructions. Each game is a sequence of levels, and a strategy that clears
one level can fail on the next, so every new level is in effect a new task. The agent records what
it learns about each game as Python scripts and text notes, while the harness keeps a complete log
of every action and observation. Our contribution is a measurement protocol that traces each
thought through these files, from the task where it forms to the task where it is corrected or
abandoned, applied to seven evaluation runs with three backbones from two model families. We find
that scripts written for one task are almost never called again in a later task (33 of 630
references cross a task boundary), because most scripts embed the state of the current level and
become invalid when the level changes. Instead, the agent rewrites its knowledge into new scripts, retyping most of
each version while keeping the general rules and dropping the level-specific details, and abandons
74\% of the scripts it wrote before a boundary. The notes, which only the model reads, are never
revised. The agent appends to them without removing earlier claims, and the contradictions that
accumulate are settled against the log. Because the log preserves everything,
the agent can discard scripts freely and reconstruct their content when needed, so it forgets
selectively, not catastrophically. The most costly error is a hard-coded value carried into a task
where it no longer holds. The fix the agent found was to turn the constant into a parameter that
must be supplied afresh for each new task. The above findings were obtained from the files the
agent wrote, without access to the model, and constitute a white-box analysis of how a coding
agent continually learns.
\end{abstract}

\section{Introduction}

Continual learning is usually framed as updating a parametric model over a sequence of tasks without
catastrophic forgetting of earlier competence, and the difficulty is attributed to a tension between
stability and plasticity \citep{parisi2019continual,delange2021survey,wang2023clsurvey}. An agent built
on a frozen foundation model, on the other hand, tackles the same continual learning problem without
weight updates: it must decide where knowledge is kept and what happens to that knowledge when the
task changes.

To study that question, we develop a coding agent whose knowledge resides entirely in a directory,
building on
the programmatic-memory design of PRO-LONG \citep{prolong2026}. The harness appends every
interaction to \texttt{logs.txt} and provides the agent only the path to that file. Any fact about
the environment must therefore be established by a script the agent writes. The memory of the agent
consists of these scripts and its notes, and both are directly readable.

We evaluate the agent on ARC-AGI-3 \citep{arcprize2026}, the interactive benchmark designed to
test general intelligence as the efficiency of acquiring skills on novel tasks
\citep{chollet2019measure}. The rules of every game must be discovered by playing it. Each
game is a sequence of levels that share one interface yet differ enough that a solution to one can
fail outright on the next (Section~\ref{sec:arc}). A level boundary is therefore a task boundary
inside a single episode, and it arrives unannounced. At each one the agent must decide what to
carry, what to rebuild and what to abandon.

Our contribution is a \textbf{measurement protocol}. Every script is assigned to the task that
was open when it was written, which yields 4{,}786 scripts across 689 boundaries
(Section~\ref{sec:measure}). The protocol needs no
access to the model that produced the memory, so it applies to any agent that writes its knowledge
to a file system. It makes retention and transfer directly countable where both were previously
inferable only from task performance.

Applying the protocol to seven deliberately varied runs across three backbones from two model
families, which clear
between 34 and 183 of the 183 available levels, we report two findings. The first is an
\textbf{empirical characterization} of cross-task update
(Sections~\ref{sec:revise}--\ref{sec:notes}). Transfer occurs by
re-derivation rather than by reference: an agent carries the rule it has learned and discards the
script that established it. The moment it writes the rule out again is the moment it separates
what the environment always does from what belonged to the task it has just left. All
backbones respond to a boundary by partitioning memory rather than by reconciling it and differ
only in the partition key.

The second finding is \textbf{negative transfer from stale memory, and its repair}
(Section~\ref{sec:fails}). A
false belief is eventually corrected by
the interaction record, so the most costly failure is a belief that is true of one task and false of
the next, which we find in three runs of one game. The one repair that works changes
how the belief is stored: the value fixed inside a script becomes an argument the script must be
given, an operation the scripts allow and the notes lack.

\section{Related Work}
\label{sec:related}

Surveys of continual learning group methods by the mechanism that protects earlier tasks while a
new one is learned: regularization of parameters, replay of stored data, and parameter isolation
\citep{parisi2019continual,delange2021survey}.
The field asks how much a learner retains, how much it can still acquire, and whether
earlier learning helps later \citep{wang2023clsurvey}. Task-free formulations drop the assumption that
boundaries are announced \citep{aljundi2019taskfree}. Regularization methods protect the
parameters that earlier tasks depend on \citep{aljundi2018mas}, and replay methods make the
question of what to keep explicit, fixing a budget and selecting into it by the gradients a sample
constrains \citep{aljundi2019gss}. Our setting inherits the question but removes the budget, because a file
system imposes none, so what to discard is decided free of the pressure of a full buffer.

Studies of continual learning in agents differ in \emph{what changes} during learning. In-context learning changes nothing persistently and is bounded by the context window. Even so,
CL-Bench \citep{asawa2026clbench} finds that keeping the full context outperforms several
purpose-built memory systems, a baseline any memory architecture must beat. Memory systems change an
external record \citep{park2023generative,packer2023memgpt}, eliminating parametric forgetting but
replacing it with a retrieval bottleneck and with consolidation error. Skill libraries change a set
of reusable procedures \citep{wang2023voyager,tang2026wikiskill}, on the premise that the collection
only grows, the premise our boundary counts contradict. Harness and loop engineering change the scaffold around a fixed model
\citep{seong2026harness,wei2026evoharness,xu2026interface,agrawal2026gepa}. These approaches
require a predefined space of admissible edits and a corpus of prior trajectories, and neither
exists on first contact with a single unseen environment. Test-time training changes weights transiently
\citep{sun2024ttt}, and memory written as code is its non-parametric counterpart: the state is a
directory of scripts and the update rule is the agent writing code.

Common to these families is that the representation in which knowledge is held is chosen in
advance rather than by the agent, and that retention and transfer are estimated from task
performance rather than read from the stored knowledge itself
\citep{delange2021survey,wang2023clsurvey}. Interpretability research
answers opacity inside the model with instruments: circuit tracing recovers a graph of the internal
computation at substantial cost per prompt \citep{ameisen2025circuit,lindsey2025biology}. In this work, by
contrast, the agent holds its state in files, which provide a complete trace of what it keeps,
revises and discards across tasks.

\section{Approach}
\label{sec:method}

\subsection{ARC-AGI-3 and the level as a task}
\label{sec:arc}

ARC-AGI-3 \citep{arcprize2026} gives an agent a $64\times 64$ grid of colored cells and seven actions named \texttt{ACTION1} to \texttt{ACTION7} together with \texttt{RESET}, and nothing else. There are no instructions, no statement of the objective, no account of what any action does and no key to the colors. The only feedback is a score that increases when a level is cleared. The agent can learn a mechanic only by acting and observing how the grid changes. 

The public demo dataset has 25 such games. A game is a sequence of levels, and clearing one loads the next
inside the same episode. The grid size, the action names and the scoring are constant across a boundary. The layout, the objects present and the winning condition change, often enough to invalidate the rules the agent learned on the previous level. Our first agent cleared level 1 (L1) of \texttt{wa30} on a theory of ``pushable boxes'' and then
failed L2 by exploiting the same theory. 
ARC Prize
gives the same warning \citep{arcmethod2026}: without a check on why a level was won, a model
carries its misconception forward.

We therefore treat each level as a task. A boundary arrives unannounced, with no signal beyond the score, and at each one the agent must decide which of the things it has learned is a law of the game and which was only a fact about the level it has left. Everything measured in Section~\ref{sec:results} happens at one of these transitions.

\subsection{Memory as prose}
\label{sec:prose}

Our first agent, a predecessor of the runs in Table~\ref{tab:iterations}, kept its memory as prose in the manner of \citet{park2023generative} and \citet{packer2023memgpt}, summarizing the interaction history and retrieving fragments of it into the prompt. That design has two weaknesses. First, summarizing discards information. Second, a rule written in prose cannot be executed, so it cannot be checked against the recorded history. Both weaknesses showed in the run. The agent compressed 2{,}416 memory entries into 106 through as many consolidation passes. Having inferred a movement budget from an on-screen bar, it filed that budget under a section titled \texttt{CONFIRMED MECHANICS} and treated it as ground truth thereafter. Had the rule been a transition function, the recorded episode could have been replayed against it. The runs studied here retain one prose component, the notes. Sections~\ref{sec:notes} and~\ref{sec:fails} show that the same pattern, correction by addition rather than by deletion, persists in them.
\subsection{Memory as executable code}
\label{sec:code}

The runs studied here keep the remainder of their memory as executable code, the scripts, whose outcomes at a task boundary are reported in Sections~\ref{sec:revise} and~\ref{sec:reuse}. A rule written as a script can be run against the recorded history, and a mismatch tells the agent that its model needs revision. The memory then accumulates as a set of artifacts (parsers, state reconstructions and simulators) that persist in a directory and can be read from outside, which is the trace this paper follows. 
\begin{figure}[t]
  \centering
  \includegraphics[width=0.84\linewidth]{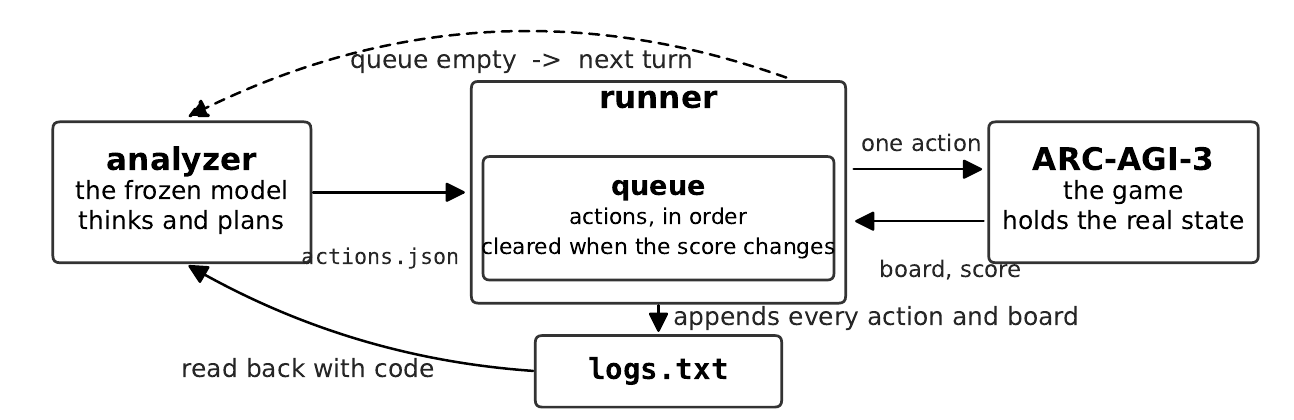}
  \caption{The loop, taken unmodified from PRO-LONG \citep{prolong2026}, with runs
  differing only in backbone, prompt and operating mode (Table~\ref{tab:iterations}). The
  analyzer, the only component that makes a decision, reads \texttt{logs.txt} and writes its
  intended actions to \texttt{actions.json}. The runner issues one action at a time, appends each
  action and the returned board to \texttt{logs.txt}, and invokes the analyzer again once its queue
  empties, discarding any queued plan when the score changes.}
  \label{fig:loop}
\end{figure}
Figure~\ref{fig:loop} shows the loop that writes this memory. Because the analyzer receives the path to the log rather
than the board itself, establishing the current or any
past state requires writing and running a script. Over an episode \texttt{logs.txt} reaches tens of megabytes, 24.9\,MB at the largest, while the prompt stays a fixed size. The analyzer therefore keeps its model of the game in its memory, the files it writes. 

A \textbf{turn} is one call to the analyzer. The model is called once, works, and emits a plan of up to the action cap, after which the runner executes the plan and calls the analyzer again. Each turn runs the analyzer as a Strands agent \citep{strands2026}, a ReAct loop \citep{yao2023react} of roughly ten tool calls that searches the log, writes a parser, runs it, inspects the output, revises and emits the plan. Six tools run inside a sandbox with no network and writes confined to the workspace of that game, so the agent can neither consult outside information nor see the files of another game. The
system prompt describes only the interface, naming the actions without saying what they do and
asserting no structure, not even that a controlled entity or an objective exists. One prompt
serves all 25 games.

Only the actions sent to the game are counted, while reasoning, tool calls and retries are not \citep{arcmethod2026}. Every action spent probing a mechanic therefore costs score, and every script run against the log is free, so the agent is rewarded for running its scripts on \texttt{logs.txt}, the record of every action already taken, rather than probing again. The setting is also close to task-free \citep{aljundi2019taskfree}, since the agent is told nothing about task identity and receives no reward during the episode, so it alone decides when to change its model.

\subsection{Measurement at a task boundary}
\label{sec:measure}

A \textbf{task} is one level of one game, and a \textbf{boundary} is the transition between two consecutive levels, the instant at which one is cleared and the next loads into the same episode.
Run R4\footnote{The implementation, prompts, tool definitions, the measurement scripts behind every count in
Section~\ref{sec:results}, and the official ARC-AGI-3 scorecard for this run will be made publicly
available.} contains 183 tasks and 158 boundaries, and
across the seven runs we observe 689, every one within a single game.

\begin{table}[t]
  \centering
  \caption{The seven runs. They differ in backbone, prompt, action cap and operating mode, and are
  not repetitions of one experiment, so we do not rank them, attribute no difference to a single
  factor, and report counts without confidence intervals. The spread lets us ask which
  behaviors hold across conditions. RHAE is Relative Human Action Efficiency as defined by ARC Prize \citep{arcmethod2026}, which rewards clearing a level in fewer
  actions than the upper-median first-time human. For reference, PRO-LONG \citep{prolong2026}
  reports 97.4\% best@2 RHAE with Fable 5 under a 2{,}000-action limit, a figure not comparable to
  R4, which differs in backbone, action cap, run selection and scoring mode. \emph{Boundaries} counts
  the within-game level transitions observed, the unit of Section~\ref{sec:results}. \emph{Scripts}
  excludes the log-reading library supplied to the Qwen runs.}
  \label{tab:iterations}
  \small
  \begin{tabular}{llrrrrr}
    \toprule
    Run & Backbone and condition & RHAE & Levels & Boundaries & Scripts & Notes \\
    \midrule
    \multicolumn{7}{l}{\emph{Anthropic backbones}} \\
    R1 & Opus 4.7, action cap 20            & 46.82\% & 116 & 107 & 898  & 225\,kB \\
    R2 & Opus 5, online mode                & 98.33\% & 181 & 158 & 675  & 1{,}375\,kB \\
    R3 & Opus 5, competition mode           & 96.87\% & 179 & 156 & 680  & 1{,}313\,kB \\
    R4 & Opus 5, prompt priors removed      & \textbf{99.95\%} & \textbf{183} & 158 & 734 & 1{,}357\,kB \\
    \addlinespace
    \multicolumn{7}{l}{\emph{Local Qwen3.8-27B-FP8 on 8 H100s, offline, action cap 50}} \\
    R5 & prompt variant \texttt{exp7}       &  8.39\% & 42 & 42 & 1{,}274 & 719\,kB \\
    R6 & prompt variant \texttt{v24}        &  6.78\% & 36 & 34 & 477  & 761\,kB \\
    R7 & prompt variant \texttt{v45}        &  6.84\% & 34 & 34 & 169  & 288\,kB \\
    \bottomrule
  \end{tabular}
\end{table}

The \textbf{memory} of the agent is the set of files it writes: its \textbf{scripts}, the Python files it
wrote and executed, and its \textbf{notes}, one or more plain-text files of beliefs per game. The harness-written record \texttt{logs.txt} is not part of the memory, because the
agent only reads it, an exclusion that matters in Section~\ref{sec:notes}.

Boundaries are recovered from the run log, which timestamps every level completion, and each
script is assigned to the task open at its write time. The console logs of the Qwen runs were not
retained, so their boundaries come from the level number stored with every action in the interaction
record. Three Qwen games write no such record and are excluded, so 94\% of Qwen scripts and all
Anthropic scripts are assigned.

\begin{table}[t]
  \centering
  \caption{The four outcomes for a script at a task boundary, how each is counted from the
  files, and where each is reported. The first two are the forms of transfer, the third is
  revision, the fourth is disposal. The outcomes apply to scripts. The notes are written for the
  model rather than for the scripts, and a correction is appended to them rather than applied, so
  their only outcome is to grow (Section~\ref{sec:notes}).}
  \label{tab:defs}
  \small
  \begin{tabular}{lp{4.1cm}p{4.1cm}l}
    \toprule
    Outcome & Counted as & What it means & Reported in \\
    \midrule
    reused by reference & a script written after the boundary imports or \texttt{exec}s it & the
      script is still imported in the next task & Section~\ref{sec:reuse} \\
    \addlinespace
    reused by copying & a different script written after the boundary contains one of its functions
      with an identical body & a function of the script is carried into the next task, the script is
      not & Section~\ref{sec:reuse} \\
    \addlinespace
    rewritten & a new version of the script is written after the boundary & the script is replaced
      by a new version in the next task & Section~\ref{sec:revise} \\
    \addlinespace
    abandoned & none of the above, no later script rewrites, imports or copies it & the script is
      not carried into the next task & Section~\ref{sec:reuse} \\
    \bottomrule
  \end{tabular}
\end{table}

A script written before a boundary has one of four outcomes after it, defined in Table~\ref{tab:defs} and each decided from the file contents alone. It is \textbf{reused by reference}, \textbf{reused by copying}, \textbf{rewritten} or \textbf{abandoned}.
Copying moves a part into another script,
rewriting replaces the script itself. In continual learning terms, reuse is forward transfer of
artifacts, rewriting and abandonment are the plastic half of the stability--plasticity trade-off. However, backward transfer is undefined because no cleared level is revisited, so the forgetting we
count is disposal of an artifact rather than loss of competence.
Reuse by reference counts both \texttt{import} and \texttt{exec} calls. Besides importing an earlier file, the agent often runs the leading part of one with \texttt{exec}, using the idiom \texttt{exec(open('cur9.py').read().split("sel=")[0])}.

\section{Results}
\label{sec:results}

\begin{figure}[t]
  \centering
  \includegraphics[width=0.74\linewidth]{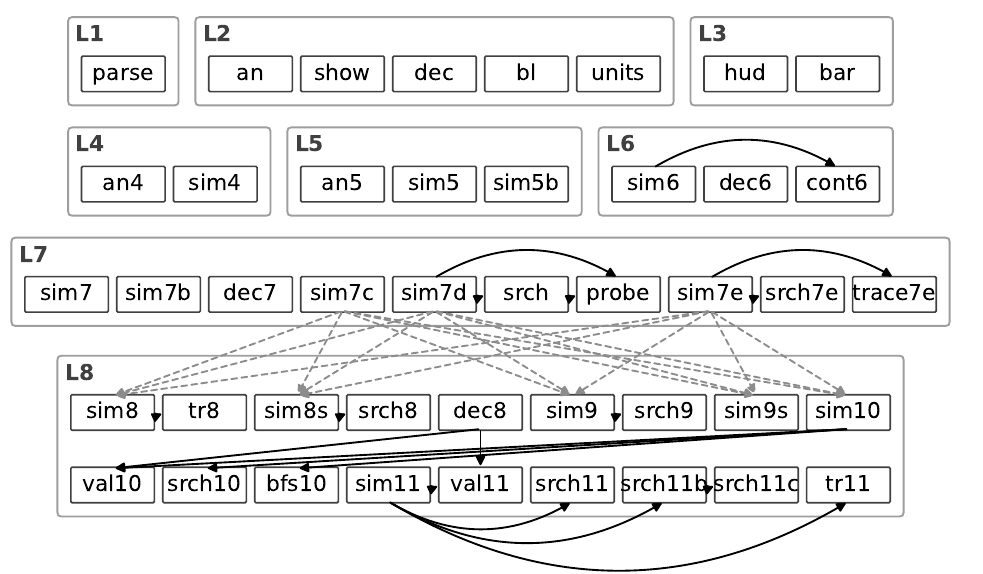}
  \caption{The reuse graph for one game, \texttt{sk48} in run R4. Each box is a script, each gray
  outline groups the scripts of one level, written left to right. A solid arrow is reuse by reference, a dashed arrow reuse by
  copying (Section~\ref{sec:measure}). \textbf{All 19 solid arrows stay inside one level.
  All 15 dashed arrows cross between levels.} Reuse also starts late: levels 1 to 6 hold 13
  scripts and one arrow, levels 7 and 8 hold 31 scripts and the other 18.}
  \label{fig:reuse}
\end{figure}

\begin{figure}[t]
  \centering
  \includegraphics[width=0.70\linewidth]{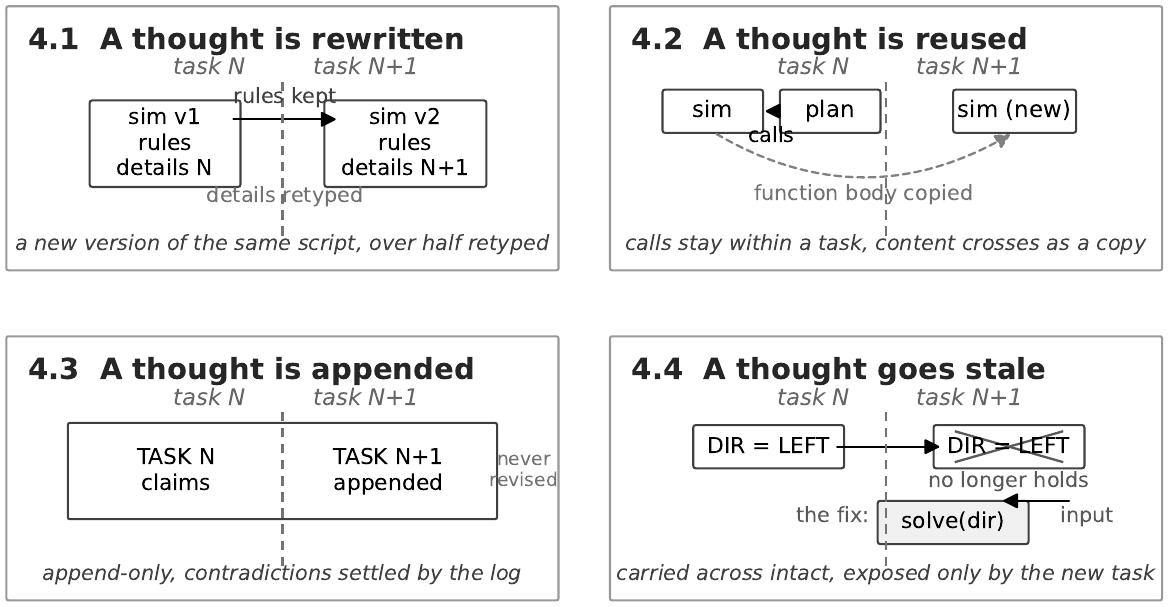}
  \caption{The four subsections of Section~\ref{sec:results}, one task change seen four ways. A thought
  is rewritten into the next version with its rules kept and its details retyped
  (Section~\ref{sec:revise}), reused by reference within the task and by copying across it
  (Section~\ref{sec:reuse}), appended to notes that only grow (Section~\ref{sec:notes}), or
  carried across stale until the new task exposes it (Section~\ref{sec:fails}). Solid arrows are actions on live scripts and the dashed arrow is content copied into a
  different script, as in Figure~\ref{fig:reuse}.}
  \label{fig:quadrant}
\end{figure}

The memory of the agent is a set of files, for each game a folder of Python scripts and its notes. A revision changes one file. The prose memory of our first agent was a single document, rewritten whole at every summary. Sections~\ref{sec:revise} and~\ref{sec:reuse} report the four script outcomes of Table~\ref{tab:defs} at the 689 boundaries, rewriting in the first and reuse by reference, reuse by copying and abandonment in the second. Section~\ref{sec:notes} turns from the scripts to the notes, which have a single outcome. Section~\ref{sec:fails} follows one belief through both kinds of memory, a value true in one task and false in the next, and finds that only the scripts offer a repair. Figure~\ref{fig:quadrant} summarizes the four subsections.

\subsection{Rewritten scripts}
\label{sec:revise}

Rewriting is frequent and its unit is a single script. In R4, 230 of the 734 scripts are a new
version of an earlier one, 101 scripts reach two or more versions and one reaches twelve. The three
Opus 5 runs write each new version as a separate script and never call the file-editing tool.
Across the seven runs, 2{,}319 scripts were written before a boundary, and 523 of them were
rewritten after it. In R4 the figures are 565 and 117.

New versions are largely retyped rather than extended. When a script is rewritten, the new version
drops 56\% of the functions of the old one in R4, 62\% and 61\% in R2 and R3, 38\% in R1, and
73\%, 74\% and 52\% in R5, R6 and R7.

What survives a rewrite is the part that does not depend on the level. The game
\texttt{su15} splits cleanly. The L6 simulator keeps five functions of the L5 simulator unaltered
and replaces four, the disc the player moves, the pieces on the board, the chasing units and the
winning condition. The five kept functions encode the rules of the game, the four replaced ones the
layout of L5. Writing a new version is the moment the agent must separate the rules of the game
from the contents of a level.

\subsection{Scripts reused by reference, reused by copying, or abandoned}
\label{sec:reuse}

Reuse by reference stays inside a level and almost never crosses one. Across the seven runs a
script imports or \texttt{exec}s another 630 times. In 597 of these cases both scripts belong to
the same level, so the reuse is within a task and is not a boundary outcome. In the remaining 33
the imported script belongs to an earlier level.

A referenced script is not a library, because it runs code the moment it is imported. Of the 273
referenced scripts, 263 do this and 229 embed five or more numbers in that code, such as the
dimensions of the current level. The L5 simulator in the R4 game \texttt{sp80} opens with \texttt{ROWS=19} and
\texttt{COLS=19}, fixing the board size as it loads. All four scripts that import it were written
during L5. L6 has a different board, so on L6 the script is invalid on import, before any of its
functions is called.

Reuse by copying is the more common route across a boundary. A function body recurs across two or more scripts 562
times, and 72 of those pairs span a level change against the 33 spanned by reference.
Figure~\ref{fig:reuse} shows both routes.

Only the three Opus 5 runs (R2--R4) reuse any script across a level change. R1 and the three Qwen
runs (R5--R7) reuse none, neither by reference nor by copying. R1 does copy 137 bodies within
levels, more than any other run, so it reuses scripts but never across a level. What the Opus 5 runs move into a later level is a copied function
rather than a retained artifact. These runs accumulate versions rather than the growing library of
reusable procedures that skill-library designs assume
\citep{wang2023voyager,tang2026wikiskill}.

Abandonment is the remainder and the most common outcome. Of the 2{,}319 scripts written before a
boundary, 1{,}726, or 74\%, are neither rewritten nor reused after it. Per run, the abandoned
share is 60\% in R3, 73\% in R2, 74\% in R4 and 88\% in R1, and 84\% in R5, 93\% in R6 and
100\% in R7. In R1, R5 and R6, which reuse nothing across a level, every script that is not
abandoned was rewritten. R7 passed few levels, so only 17 of its 169 scripts were written in a level
the agent later passed, and none of the 17 was rewritten.

\subsection{Updates to the notes}
\label{sec:notes}

\begin{table}[t]
  \centering
  \caption{How the notes of each run accommodate a task change. \emph{Notes} totals every file matching
  \texttt{notes*.md}, which matters for R1 because it writes 101 of them. \emph{Under a level
  heading} is the share of notes text beneath an explicit level heading. \emph{Level-named files}
  counts scripts whose filename names a level. \emph{Index} is the key under which the notes of
  each run are organized, and the three backbones use three different ones.}
  \label{tab:notes}
  \small
  \begin{tabular}{llrrrrr}
    \toprule
    Run & Index & Notes & Level & Action or turn & Under a level & Level-named \\
        &       & (kB)  & headings & headings     & heading       & files \\
    \midrule
    R1 & one file per level       & 225      & 89  & 57  & 59\% & 227 \\
    R2 & level sections, one file & 1{,}375  & 552 & 806 & 93\% & 47 \\
    R3 & level sections, one file & 1{,}313  & 414 & 754 & 93\% & 68 \\
    R4 & level sections, one file & 1{,}357  & 397 & 824 & \textbf{98\%} & 104 \\
    \addlinespace
    R5 & its own turn             & 719      & 88  & 292 & 55\% & 184 \\
    R6 & its own turn             & 761      & 87  & 379 & 48\% & 44 \\
    R7 & its own turn             & 288      & 59  & 183 & 57\% & 28 \\
    \bottomrule
  \end{tabular}
\end{table}

The notes have a single outcome, growth. The two parts of the memory have different readers. Scripts
are written to compute facts from the record, and 1{,}979 of the 4{,}907 open \texttt{logs.txt}.
The notes are written for the model and have no other reader. A claim in the notes is therefore not
checked by execution directly, and a correction is appended rather than applied, so it takes effect only
when the model re-reads the file.

The notes therefore need an index, and the seven runs adopt three, listed in the \emph{Index}
column of Table~\ref{tab:notes}. The Opus 5 runs keep a single file per game and index it by level, placing 93 to
98\% of the text under a level heading. Opus 4.7 indexes by file, writing one notes file per level
(\texttt{notes\_lvl2.md} through \texttt{notes\_lvl8.md} in \texttt{ar25}) and naming a level in 227 of its 898 script filenames. The Qwen runs index by
their own turn. In R5, 73\% of headings name a turn and 13\% a level, against 12\% and 27\% in R4.

A corrected claim is never removed. In the R4 game \texttt{wa30} the notes first describe a
two-cell model of the player. A later section titled \texttt{KEY MECHANIC CORRECTION} declares it
wrong and replaces it with a one-cell model. Eighty actions later a section titled \texttt{MAJOR
REVISION} restores the two-cell model. The first model was right and the correction wrong, yet both
remain in the file 65 lines apart, and neither is marked as withdrawn. In \texttt{sb26}, R2 and R3 each
add a section titled \texttt{CORRECTION} and leave the false sentence standing above it. The agent
works around a contradiction rather than settling it, which is workable because \texttt{logs.txt}
lies outside the memory and decides between the claims the memory cannot reconcile. The notes are a
dated record of past beliefs rather than a statement of current ones, which is what makes them
traceable. Every thought the agent held about a game, including the ones it later abandoned, is
still there in the order it was formed.

\subsection{Negative transfer from stale memory}
\label{sec:fails}

The failure that recurs in both model families is negative transfer from stale memory, a belief
that is true in one task and false in the next. The log of the first task cannot expose it, because
it is true there. Both kinds of memory carry it into the next task, each rewrite by retyping the
constant into the new version (Section~\ref{sec:revise}) and the notes by restating it
(Section~\ref{sec:notes}).

The clearest case is the game \texttt{m0r0}, whose horizontal control is mirrored. The action
that brings the two markers together in L4 is reversed in L5 and reversed again at the next
boundary. All three Opus 5 runs carry the L4 direction into L5. They differ in the repair, which in
every run is made in the scripts. Two runs changed the form of the belief, replacing the fixed
value with a parameter (\texttt{def solve(polarity, ...)} in one and \texttt{S3 =
int(sys.argv[1])} in the other), so the direction must be supplied at run time. R4 kept the value in
a dictionary and edited it. It carried the wrong value into L5, corrected it there, carried the
correction into L6 where it was wrong again, and restored the original. The parameter is the repair
that works. At L6, R2 and R3 tried both values and settled the direction within a few actions,
whereas R4 carried its L5 value across and failed again. The notes of all three runs restate the direction as a fact under each level heading, and at L5
record the reversal as a correction, but none of them records the direction as a quantity to be
re-established at every level. That representation exists only in the scripts, as the parameter.

The same failure appears in the Qwen runs, and in R6 the repair runs backwards. In the game
\texttt{tu93} the L2 simulator takes the step size as an argument, \texttt{def move(r, c, mv,
n=2, ...)}. The L3 rewrite replaces the function with a table of fixed offsets,
\texttt{DIRS=\{'ACTION1':(-2,0), 'ACTION2':(2,0), ...\}}, so the value that L2 had left open is
closed again. The pattern holds across R6, whose scripts hold almost one hard-coded number per
line of code. The cases in both families share one structure. A value that changes from level to
level is written into the script as a constant, and the repair, where one was made, turns that
constant into a parameter.

\section{Conclusion}

We traced how a coding agent learns across the abstract reasoning tasks of ARC-AGI-3 by reading
the scripts and notes it writes and the log the harness keeps, and the same outcomes appeared, in
different proportions, in all three backbones. The agent reuses scripts within a task, and across
tasks it rewrites a few, copies a function from fewer, and abandons most, keeping the general rules
and dropping the level-specific details. The notes only grow, and the log settles what they cannot.
In the taxonomy of continual learning this is the isolation strategy \citep{delange2021survey},
adopted without instruction. The agent discards only beliefs, never evidence, so a wrong disposal
costs a re-derivation rather than a competence, and forgetting is selective rather than
catastrophic. The most costly error is a hard-coded value carried into a task where it no longer
holds, and the fix the agent found, turning the constant into a parameter, is as visible in the
trace as the error. Learning in this agent takes place in its memory, proceeds by selecting what to
keep and what to discard, and every step of it leaves thoughts that can be traced, making the
continual learning of a coding agent transparent from the first task to the last.

\bibliographystyle{plainnat}
\bibliography{refs}

\end{document}